\documentclass[letterpaper, 10 pt, conference]{ieeeconf}  
\IEEEoverridecommandlockouts 
\title{\LARGE \bf
Uncertainty Estimation for 3D Dense Prediction \\ via Cross-Point Embeddings}

\author{Kaiwen Cai\authorrefmark{1}, 
Chris Xiaoxuan Lu\authorrefmark{2}, 
Xiaowei Huang\authorrefmark{1}\\
\authorrefmark{1}University of Liverpool, \authorrefmark{2}University of Edinburgh
}

\usepackage{geometry}
\usepackage{color}

\definecolor{RightColor}{RGB}{39, 174, 96} 
\definecolor{WrongColor}{RGB}{202, 62, 71} 
\definecolor{QueryColor}{RGB}{155, 89, 182} 

\usepackage{amsmath}
\usepackage{amsfonts}
\usepackage{amssymb}

\usepackage{dsfont}

\usepackage{bbm}

\usepackage{siunitx}

\usepackage{graphicx}
\graphicspath{ {./figures/} }

\usepackage{booktabs}
\usepackage{float}  
\usepackage{threeparttable} 
\usepackage{tablefootnote}  
\usepackage{multirow}
\usepackage{colortbl}

\usepackage{cite}
\usepackage{color}
\usepackage[dvipsnames]{xcolor} 

\usepackage[commentmarkup=uwave]{changes} 
\definechangesauthor[name={kw}, color=Maroon]{KW}
\definechangesauthor[name={xiaowei}, color=blue]{xiaowei}
\definechangesauthor[name={CHRIS}, color=magenta]{CHRIS}

\newcommand{\sysname}{{CUE}}   

\newcommand{\rev}[1]{{\color{black}{#1}}}    
\newcommand{\revised}[1]{{\color{black}{#1}}}    
\renewcommand{\thetable}{\Roman{table}} 
\makeatletter
\let\NAT@parse\undefined
\makeatother
\usepackage[colorlinks, linkcolor=black, anchorcolor=black, citecolor=black]{hyperref}
\begin{document}
\maketitle
\thispagestyle{empty}
\pagestyle{empty}

\begin{abstract}
    Dense prediction 
    tasks are common for 3D point clouds, but the uncertainties \revised{inherent} in massive points and their embeddings have long been ignored. In this work, we present \sysname, a novel uncertainty estimation method for dense prediction tasks \rev{in} 3D point clouds. Inspired by metric learning, the key idea of \sysname\ is to explore cross-point embeddings upon a conventional \rev{3D} dense prediction pipeline. Specifically, \sysname\ involves building a probabilistic embedding model and then enforcing metric alignments of massive points in the embedding space. \rev{We also propose \sysname+, which enhances \sysname\ by explicitly modeling cross-point dependencies in the covariance matrix.} We demonstrate that both \sysname\ and \sysname+ are generic and effective for uncertainty estimation in 3D point clouds \rev{with} two different tasks: (1) in 3D geometric feature learning we for the first time obtain well-calibrated uncertainty, and (2) in semantic segmentation we reduce uncertainty's Expected Calibration Error of the state-of-the-arts by \revised{16.5\%}. All uncertainties are estimated without compromising
    predictive performance.
 \end{abstract}


\section{Introduction}

The process of predicting a label of each point in a point cloud is known as 3D dense prediction. \revised{It is a crucial aspect of} robotic perception and autonomy, enabling tasks such as semantic segmentation, depth completion, and scene flow estimation.


UNet-based networks have \revised{become} the de-facto choice for point cloud dense prediction \cite{ronneberger2015u, choy2019fully, ao2021spinnet, thomas2019kpconv}. In a UNet-like network, 
the input and the output of two correspondingly linked layers have the same number of points, e.g., if the input point cloud is \revised{represented} by a $N \times 3$ tensor, then the output of its correspondingly linked layer is a $N \times D$ tensor. In this regard, the output can also be \revised{considered} as an embedding map, and a dense prediction network can then be decomposed as an embedding learning network and a task-specific regressor (or classifier).
Thus, 
the \revised{core} of the dense prediction task is embedding learning.

Embedding learning aims to learn a discriminative model that \revised{maps samples of the same class closer together and those of different classes farther apart in the embedding space.}
Successful embedding learning \revised{facilitates} many downstream tasks, including image retrieval \cite{musgrave2020metric}, face recognition \cite{meng2021magface} and zero-shot learning \cite{bucher2016improving}. 
In addition to enhancing the discriminative capability of the embedding model, quantifying its uncertainty is also \revised{gaining significant attention}.

\begin{figure}[htbp]
    \centering
    \includegraphics[width=3.2in]{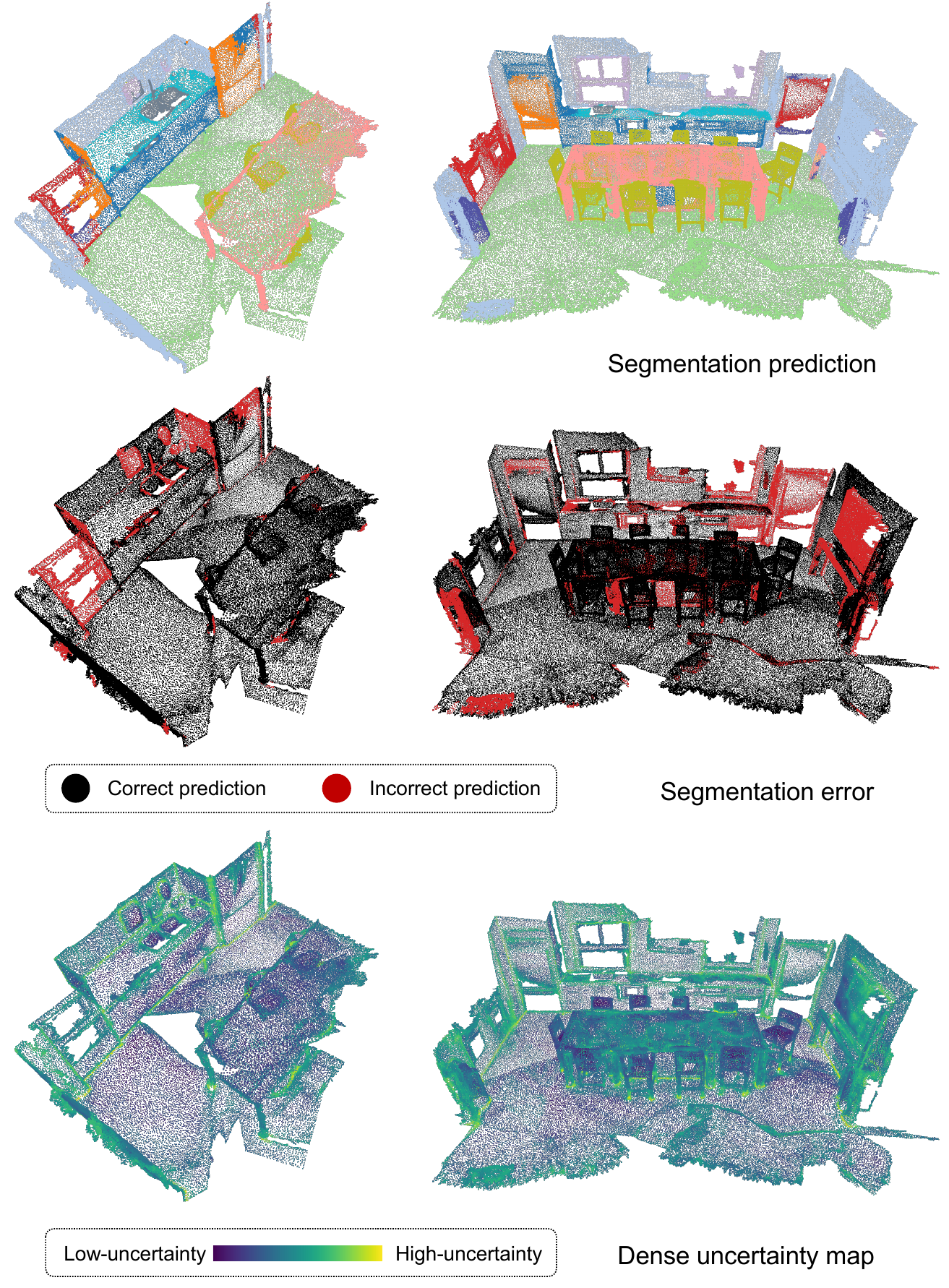} 
    \caption{In a \rev{3D} dense prediction task, i.e., 3D semantic segmentation, we present the segmentation prediction (top), segmentation error  (middle) and dense uncertainty map (bottom, estimated by \sysname+) of two scenes from ScanNet validation split. Incorrect predictions tend to have high uncertainties.} 
    \label{fig_qua_scannet}
    \vspace{-0.75cm}
\end{figure}

For dense prediction tasks of point clouds, it is \revised{beneficial} that an uncertainty level could be provided in conjunction with the point-wise labels to make its downstream decision-making more information-aware. 
\revised{For example, in an autonomous vehicle scenario where semantic labels of each point on the road are predicted, an estimated uncertainty level would aid the vehicle in determining when to trust the prediction and optimize planning and control.}
Such promising benefits have stimulated the development of various uncertainty estimation methods for different dense prediction tasks, including
 (1) using the output of the logit layer to calculate softmax entropy~\cite{czolbe2021segmentation}, (2) building a two-head network to predict the mean and variance of an embedding separately \cite{kendall2017uncertainties}, and (3) resorting to a Bayesian Neural Network (BNN) model with Monte Carlo Dropout (MCD) to approximate posterior weights \cite{qi2021neighborhood}. 

However, two major issues \revised{persist in 
current methods for estimating uncertainty in the dense prediction of 3D point clouds}.
\revised{Firstly}, points can only interact within the limited receptive field of convolution kernels and \revised{require a shared MLP to facilitate implicit interactions among logits}.
\revised{This} under-treatment of cross-point dependencies, unfortunately, often results in suboptimal uncertainty estimation as evidenced by~\cite{monteiro2020stochastic}. 
\revised{Secondly}, 
a notable trait of the 
\revised{prevalent}
dense prediction networks is that they are sequential compositions of embedding learning networks and task-specific regressors (or classifiers). While prior arts have shown that 
\revised{incorporating} embedding learning in regression or classification tasks
can yield better predictive performance \cite{li2021learning,wang2021exploring},
it is largely under-explored 
if utilizing embedding learning  
can also give rise to better-calibrated uncertainty. 


In this paper, we propose a novel and generic uncertainty estimation pipeline, called \sysname\ in the paper for \textbf{C}ross-point embedding \textbf{U}ncertainty \textbf{E}stimation, to bridge the gap between the dense prediction of point clouds and its uncertainty quantification. \sysname\ involves \revised{constructing} a probabilistic embedding model and enforcing metric alignments of massive points in the embedding space. To \revised{address} the aforementioned issues, \sysname\ \revised{emphasizes} the importance of 
embedding learning, and exploits this embedding space through a diagonal multivariate Gaussian model 
that facilitates
cross-point interactions. \revised{Additionally}, we propose \sysname+ that further utilizes cross-point dependencies by a low-rank multivariate Gaussian model. \revised{The} low-rank covariance matrix in \sysname+ explicitly \revised{captures} off-diagonal elements' dependencies while maintaining computational efficiency. 
Our specific contributions are:
\begin{itemize}
    \item 
    \revised{We propose a generic framework for estimating uncertainty in dense prediction tasks of 3D point clouds}.
    \item 
    We propose a novel approach that fully \revised{leverages} cross-point information for \revised{estimating uncertainty}.
    \item We validate our proposed method on two representative dense prediction tasks, with the experimental results consistently \revised{demonstrating} that our method produces better-calibrated uncertainty than state-of-the-art \revised{methods} without \revised{compromising} predictive performance. 
    \item Source code of both \sysname\ and \sysname+ is available at: \\\url{https://github.com/ramdrop/cue}. 
\end{itemize}

\section{Related Work}


\subsection{Dense Prediction of 3D Point Cloud}

\revised{Given} the dense nature of the 3D point cloud, we focus on dense prediction tasks, e.g., 3D geometric feature learning and 3D semantic segmentation.

\textit{3D Geometric Features Learning}: To find the correspondences \revised{between point clouds} in the absence of relative transformation information, \rev{a number of methods are} to convert point clouds from the 3D Euclidean space to a feature space, where correspondences are \revised{identified as} the nearest neighbors. Early work \revised{rely} on hand-crafted features \cite{salti2014shot, rusu2009fast} \revised{to perform this conversion}.
Recently\revised{,} deep learned geometric features are becoming \revised{increasingly} popular, which are \rev{typically} based on volumetric and \rev{pointwise} operations on point clouds: (1) \rev{Volumetric methods such as 3DMatch \cite{zeng20173dmatch} and FCGF \cite{choy2019fully} learn patch and point descriptors respectively by applying a 3D Convolutional Neural Network (CNN) on volumetric input.} 
(2) \rev{Point-wise methods such as} PointNet \cite{qi2017pointnet} use parallel shared MLP to learn global or dense features, and DGCNN \cite{wang2019dynamic} combines pointwise MLP with dynamic graph neural networks \rev{to obtain} flexible and effective feature extractors for unordered point clouds. 
Nevertheless, \rev{these methods primarily} focus on improving predictive performance while ignoring the uncertainty inherent in massive points.
\begin{figure}[htbp]
    \centering
    \includegraphics[width=3.3in]{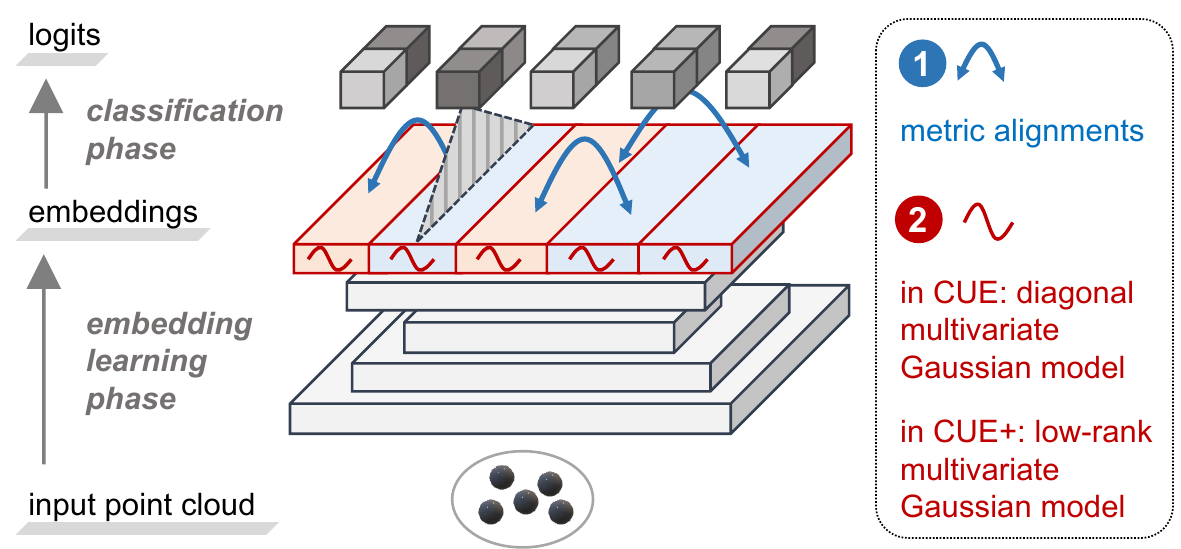} 
    \caption{\revised{An overview of the proposed \sysname\ and \sysname+. We take semantic segmentation for instance where there are $5$ points in the input point cloud and $2$ classes in the labels. \sysname\ explores cross-point embeddings by building a probabilistic embedding model (Red curves) and enforcing metric alignments (Blue arrows), and \sysname+ goes further by replacing the diagonal covariance matrix with a low-rank covariance matrix.}} 
    \label{fig_pipeline2}
    \vspace{-0.5cm}
\end{figure}

\textit{3D Semantic Segmentation}:
PointNet \cite{qi2017pointnet} is the very first work for 3D point cloud learning, while its shared-MLP architecture shows strong representation capability,
it ignores the local context. Following works propose different solutions to make for this limitation: PointNet++ \cite{qi2017pointnet++} adopts hierarchical sampling strategies, KPConv \cite{thomas2019kpconv} proposes a kernel-based MLP operation mimicking convolution, \revised{MinkowskiNet \cite{choy20194d} extends 3D convolution to 4D space} and \rev{develops} sparse operation library for point clouds, and PointTransformer \cite{zhao2021point} shows the power of Transformer mechanisms in point cloud processing.

\vspace{-0.1cm}
\subsection{\rev{Uncertainty Estimation}}
\textit{Embedding Learning Uncertainty}:
Kendall \cite{kendall2017uncertainties} categorizes uncertainties in deep learning as two types: aleatoric uncertainty and epistemic uncertainty. Aleatoric uncertainty \rev{arises} from data noises, while epistemic uncertainty refers to model uncertainty which can be reduced with sufficient training data. Embedding learning is \rev{commonly} applied to image recognition tasks, where most methods focus on estimating aleatoric uncertainty. \rev{For example,} PFE \cite{shi2019probabilistic} models face embeddings as Gaussian distributions and uses the proposed Mutual Likelihood Score to measure the likelihood of two embeddings belonging to the same class. DUL \cite{chang2020data} proposes to learn aleatoric uncertainty for both regression and classification face recognition tasks. BTL \cite{warburg2021bayesian} proposes a Bayesian loss to learn aleatoric uncertainty in place recognition. RUL \cite{zhang2021relative} uses relative uncertainty measurements to learn aleatoric uncertainty. 

\rev{The image recognition tasks discussed above involves learning a single feature for an entire image, however, in the dense prediction task of point clouds, thousands of features (i.e., equals to the number of points in the point cloud) need to be learned for a single point cloud.} Furthermore, image recognition is applied to regular-size images, while point clouds are totally unordered and have a varied size. The large number of features within a batch and irregular input size render it rather challenging to estimate uncertainty for a 3D point cloud.

\textit{Semantic Segmentation Uncertainty}:
Popular uncertainty estimation methods for semantic segmentation, as outlined in \cite{jungo2019assessing},  include softmax entropy \cite{czolbe2021segmentation}
, BNN \cite{kendall2015bayesian}, 
learned aleatoric uncertainty \cite{kendall2017uncertainties}
, auxiliary network \cite{zheng2021rectifying} 
and variance propagation based on Assumed Density Function (ADF) \cite{cortinhal2020salsanext}.  
\rev{While these approaches are widely used, they often assume independence between pixels or points. 
This lack of consideration for cross-pixel or cross-point dependencies can lead to less accurate uncertainty estimation \cite{monteiro2020stochastic}.}



Embedding learning has been explored in \rev{in the context of} image segmentation \rev{with studies such as} \cite{wang2021exploring} and \cite{tang2022contrastive} \rev{demonstrating that} contrastive learning can optimize embedding space and improve prediction performance in semantic segmentation tasks. \rev{Research like} \cite{li2021learning} also \rev{supports the idea} that optimized embeddings can contribute to improved predictive performance. 
However, \rev{these studies primarily focus on using embedding learning to enhance prediction performance}, rather than for estimating uncertainty. 
SSN \cite{monteiro2020stochastic} utilizes a low-rank multivariate Gaussian model to account for cross-pixel dependencies, but it is developed for logits and does not \rev{incorporate} embedding optimization.

The proposed \sysname\
is based on a probabilistic embedding model and enforces metric alignments in the embedding space by using bayesian triplet loss.
Bayesian triplet loss has been used in \cite{warburg2021bayesian} \rev{for} image recognition, but with some key differences.
\rev{Firstly}, the image recognition \cite{warburg2021bayesian} requires a single embedding for an entire image (i.e., a set of all pixels), while massive point-wise embeddings are desired in \sysname. \revised{This means that while traditional 2D CNNs can be used to extract image features, an efficient and effective network for extracting both point features and uncertainties remains unknown. Thus, we investigate efficient networks and propose sampling strategies for utilizing cross-point embeddings of massive and unordered points in a batch;} \rev{Secondly}, the probabilistic embedding model of \cite{warburg2021bayesian} ignores the cross-point dependencies, \rev{whereas the proposed \sysname+ addresses} this issue by using a low-rank multivariate Gaussian model.

\section{Method}

\subsection{\rev{Probabilistic Embedding Model}}
A dense prediction network maps a batch of points to a set of scalars. \rev{This} process can be \rev{broken down} into a metric learning phase and a task-oriented regression or classification phase. \rev{Specifically,} 
given a point cloud $\boldsymbol{\mathcal{P}} \in \mathbb{R}^{N \times 3}$, where $N$ is the number of points, the network $f_{\theta}$ maps it to a set of embeddings $\boldsymbol{\mathcal{X}} \in \mathbb{R}^{N \times D}$, where $D$ is the embedding dimension. The metric learning phase can be written as $\boldsymbol{\mathcal{X}} = f_{\theta}(\boldsymbol{\mathcal{P}})$. 
\rev{Then,} 
a task-oriented regressor (or classifier) $f_{r}$ generates predictions $\boldsymbol{\mathcal{Y}} \in \mathbb{R}^{N \times 1}$ (or $\boldsymbol{\mathcal{Y}} \in \mathbb{R}^{N \times C}$ where $C$ is the number of class )
for the set of embedding $\boldsymbol{\mathcal{X}}$. This regression or classification phase can be written as $\boldsymbol{\mathcal{Y}} = f_{r}(\boldsymbol{\mathcal{X}})$. 
 In the above formulation, predictions are \rev{treated} as deterministic \rev{and do not take into account the inherent noise from the data.} 
\rev{In comparision}, a probabilistic prediction model (e.g., probabilistic semantic segmentation~\cite{kendall2017uncertainties}) \rev{represents} the predictions as a Gaussian distribution, which provides uncertainty level \rev{in addition to} the prediction. 
\rev{However,} the embeddings are still \rev{treated} deterministic and \rev{are given equal weight}, meaning each embedding contributes equally to the regressor (or classifier).

Inspired by \rev{the use of} probabilistic contrastive learning in face recognition \cite{shi2019probabilistic,chang2020data}, we \rev{build} a probabilistic embedding model for a point cloud, with embeddings represented by a diagonal multivariate Gaussian distribution, \rev{which can be written as} 
\begin{equation}
    \label{eq_distribution}
    \boldsymbol{\mathcal{X}} \sim \boldsymbol{\mathcal{N}}\left ( \boldsymbol{\mu}, \boldsymbol{\Lambda}\right ),
\end{equation}
where $\boldsymbol{\mu}=f_\mu(\boldsymbol{\mathcal{P}})\in \mathbb{R}^{N \times D}$ and $\boldsymbol{\Lambda}=f_\sigma(\boldsymbol{\mathcal{P}}) \in \mathbb{R}^{(N \times D) \times (N \times D)}$ 
is a diagonal matrix. $f_\mu$ and $f_\sigma$ representss the mean branch and variance branch of the network $f_\theta$, respectively. We will later propose a full-covariance multivariate Gaussian model and \rev{demonstrate} its superiority in Sec. \ref{lrmg}.

\revised{\subsection{Metric Alignments of Embeddings}}



\rev{Once the probabilistic embedding model has been constructed, the next step is to optimize the embedding space and obtain the uncertainty.}

\revised{A traditional probabilistic prediction pipeline \cite{kendall2017uncertainties} only 
allows for implicit interactions between logits through a shared multi-layer perceptron (MLP), 
i.e., there is no explicit interaction within layers.} In contrast, the above probabilistic embedding model 
\rev{allows for 
enhanced} interactions of logits and the estimatimation of  uncertainties.
An overview of the proposed \sysname\ and \sysname+ is presented in Fig. \ref{fig_pipeline2}, where \sysname\ explores cross-point embeddings by constructing a probabilistic embedding model and enforcing metric alignments, and \sysname+ goes further by using a diagonal covariance matrix \rev{in place of} a low-rank covariance matrix. \rev{In the following discussion}, we will first describe \sysname\ which is based on the diagonal multivariate Gaussian model \rev{and then present} an improved version, \sysname+, which is based on the low-rank multivariate Gaussian model. 

\subsubsection{\sysname}
Given a triplet \rev{of samples,} $\{\boldsymbol{P_a},\boldsymbol{P_p}, \boldsymbol{P_n} \vert \boldsymbol{P_i} \in \mathbb{R}^{1 \times 3}, i=a, p, n\}$, their embeddings are obtained as  $\{\boldsymbol{X_a},\boldsymbol{X_p}, \boldsymbol{X_n} \vert \boldsymbol{X_i} \sim \boldsymbol{\mathcal{N}}(\boldsymbol{\mu}_i, \boldsymbol{\Sigma}_i), \boldsymbol{\mu}_i \in \mathbb{R}^{1 \times D}, \boldsymbol{\Sigma}_i \in \mathbb{R}^{1 \times D}, i=a, p, n\}$, where the subscripts $a, p, n$ denote an anchor, positive and negative sample, respectively. In the probabilistic setting, we are interested in the probability of the positive embedding being closer to the anchor \rev{than the negative one}:
\vspace{-0.1cm}
\begin{equation}
    \label{eq_bayesian_triplet}
P(\Vert \boldsymbol{X}_a - \boldsymbol{X}_p \Vert - \Vert \boldsymbol{X}_a - \boldsymbol{X}_n \Vert + m < 0).
\end{equation}
\vspace{-0.2cm}
Rewrite it as
\begin{equation}
    \label{eq_normal}
    P(\tau<-m),
\end{equation}
where the new distribution \revised{$\tau= \sum_{d=1}^D \boldsymbol{T} ^d=\sum_{d=1}^D (\boldsymbol{X}_a^d - \boldsymbol{X}_p^d)^2 - (\boldsymbol{X}_a^d - \boldsymbol{X}_n^d)^2$}, and $d$ \rev{denotes} $d^{th}$ dimension . According to central limit theorem, $\tau$ will approximate a normal distribution when $D$ is large, i.e.,$\frac{\tau - \mu_{\tau}}{\sigma_{\tau}} \thicksim \mathcal{N}(0, 1)$, where $\mu_\tau$ and $\sigma^2_\tau$ are the mean and the variance of the distribution $\tau$. Then \revised{(\ref{eq_normal})} is solved as $P(\tau<-m) = \Phi_{\mathcal{N}(0, 1)} (\frac{-m - \mu_{\tau}}{\sigma_{\tau}})$,
where $\Phi$ is the Conditional Density Function (CDF). \rev{The task now} is to find an analytical solution of $\mu_\tau$ and $\sigma_\tau$.
The mean $\mathbb{E}^\prime [\tau]$ and variance $\mathbb{D}^\prime[\tau]$ of a single dimension is given as (the superscript $d$ at the right-hand side is omitted for brevity)
\begin{equation}
\begin{aligned}
\mathbb{E}[\boldsymbol{T}^d] &= \mu_{p}^{2}+\sigma_{p}^{2}-\mu_{n}^{2}-\sigma_{n}^{2}-2 \mu_{a}\left(\mu_{p}-\mu_{n}\right), \\
\mathbb{D}[\boldsymbol{T}^d] &= 2[\sigma_{p}^{4}+2 \mu_{p}^{2} \sigma_{p}^{2}+2\left(\sigma_{a}^{2}+\mu_{a}^{2}\right)\left(\sigma_{p}^{2}+\mu_{p}^{2}\right)- 2 \mu_{a}^{2} \mu_{p}^{2}\\&-4 \mu_{a} \mu_{p} \sigma_{p}^{2}] + 2[\sigma_{n}^{4}+2 \mu_{n}^{2} \sigma_{n}^{2}+2\left(\sigma_{a}^{2}+\mu_{a}^{2}\right)\left(\sigma_{n}^{2}+\mu_{n}^{2}\right)\\&-2 \mu_{a}^{2} \mu_{n}^{2}-4 \mu_{a} \mu_{n} \sigma_{n}^{2}] -8 \mu_{p} \mu_{n} \sigma_{a}^{2}.
\end{aligned}
\end{equation}
Since the embedding model is assumed to be isotropic, we arrive at
\begin{equation}
    \mu_\tau = \sum_d^D \mathbb{E}[\boldsymbol{T} ^d], \quad \sigma _\tau^2 = \sum_d^D \mathbb{D}[\boldsymbol{T} ^d].
\end{equation}
\vspace{-0.3cm}

In summary, after the network generates a set of embeddings for a point cloud, we calculate the probability of the positive embedding being closer to the anchor than the negative one, and the goal of training is to minimize the metric loss derived from \revised{(\ref{eq_bayesian_triplet})}:
\revised{
\begin{equation}
    \label{eq_metric_loss}
    \begin{aligned}
&L_{M} \\&= -\frac{1}{T} \sum _{t=1}^T \log{P(\Vert \boldsymbol{X}_{t,a} - \boldsymbol{X}_{t,p} \Vert - \Vert \boldsymbol{X}_{t,a} - \boldsymbol{X}_{t,n} \Vert <-m )}
    \end{aligned}
\end{equation}
}
where $T$ is the number of total triplets in a mini-batch.

\subsubsection{\sysname+}
\label{lrmg}

\begin{figure}[t]
    \centering
    \includegraphics[width=3.3in]{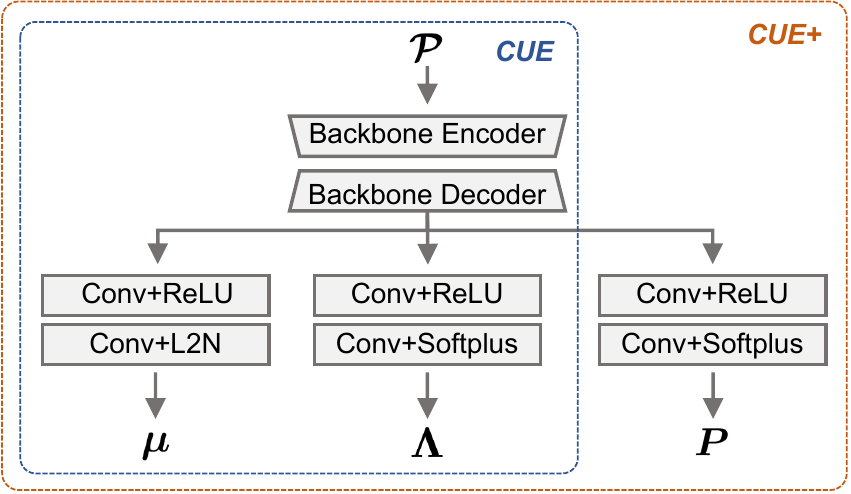} 
    \caption{The network architectures of \sysname\ and \sysname+: $\boldsymbol{\mathcal{P}}$ \rev{represents} a 3D point cloud, $\boldsymbol{\mu}$ the embeddings' mean, $\boldsymbol{\Lambda}$ \rev{the} diagonal elements of embeddings' covariance matrix, $\boldsymbol{P}$ \rev{the} scale factor of embeddings' covariance matrix.} 
    \label{fig_network}
    \vspace{-0.6cm}
\end{figure}

Points usually \rev{have a} spatial correlation with their neighbors. For example, points at the boundaries of an object usually exhibit high uncertainty since the points around the boundary have varied semantic labels.
However, \sysname\ fails to model point-wise dependencies because the diagonal covariance matrix of \sysname\ (see \revised{(\ref{eq_distribution})}) is based on the assumption that points are independent of each other. To \rev{address} this issue, we propose \rev{a solution to further capture} the point-wise dependencies by a full-covariance multivariate Gaussian model. Specifically, the diagonal covariance matrix in \revised{(\ref{eq_distribution})} is replaced with a full covariance matrix $\boldsymbol{\Sigma} \in \mathbb{R}^{(N \times D) \times (N \times D)}$
\begin{equation}
    \boldsymbol{\mathcal{X}} \sim \boldsymbol{\mathcal{N}}\left ( \boldsymbol{\mu}, \boldsymbol{\Sigma}\right ),    
\end{equation}
where $\boldsymbol{\mu} \in \mathbb{R}^{N \times D}$. However, the computational complexity of the full covariance matrix  $\boldsymbol{\Sigma}$ scales with the square of $N$, and a point cloud usually consists of tens of thousands of points, i.e., $N>10^4$. This makes training networks difficult. To alleviate this issue, we resort to a low-rank parameterization of the covariance matrix \cite{magdon2010approximating}
\begin{equation}
    \boldsymbol{\Sigma} = \boldsymbol{P}\boldsymbol{P}^T+\boldsymbol{\Lambda}, 
\end{equation}
where the scale factor $\boldsymbol{P} \in \mathbb{R}^{(N\times D)\times K}$ and $K$ is the rank of the parameterization, $\boldsymbol{\Lambda} \in \mathbb{R}^{(N\times D) \times (N\times D)}$ and $\boldsymbol{\Lambda}$ is a diagonal matrix. The pipeline based on a low-rank covariance matrix \rev{is referred to} as \sysname+ as it goes beyond \sysname\ by learning parameters of additional elements other than diagonal elements of the covariance matrix. 
\rev{This allows for explicit description of point-wise dependencies through the learned variances, making it a more powerful model than CUE.}

For ease of application, we \revised{follow \cite{magdon2010approximating} and} choose $K=1$. 
$L_M$ is then used to train the network. By experimental results, we show that \sysname+ generates better-calibrated uncertainty than \sysname\ (see Sec. \ref{experiments}).

The network architectures of the proposed \sysname\ and \sysname+ are shown in Fig. \ref{fig_network}, where $\boldsymbol{\mathcal{P}}$ \rev{represents} a 3D point cloud. The backbone encoder and decoder can be chosen from any UNet-like network. We add three branches to predict the mean $\boldsymbol{\mu}$, \rev{the} diagonal covariance matrix $\boldsymbol{\Lambda}$ and the scale factor $\boldsymbol{P}$. \revised{As L2-Normalization has been shown to improve prediction performance\cite{choy2019fully, arandjelovic2016netvlad}, we add an L2-Normalization layer to the end of the $\boldsymbol{\mu}$ branch. For the variance branch, we follow the practice of \cite{warburg2021bayesian} and add softplus layers at the end of the $\boldsymbol{\Lambda}$ and $\boldsymbol{P}$ branches.}

\revised{\subsection{Cross-point Embedding Sampling}

In the previous section, we discussed two methods to enforce metric alignments on the triplets of point embeddings. In this section, we will explain how these triplets are constructed.

In 3D dense prediction tasks, we categorize cross-point embeddings into two types: 1) \underline{C}ross-point \underline{E}mbeddings within a \underline{S}ingle point cloud (CES) and 2) \underline{C}ross-point \underline{E}mbeddings among \underline{M}ultiple point clouds (CEM). 
Depending on the specific 3D dense prediction tasks, we explore different types of embeddings as follows.

\subsubsection{CES} \label{sec_ces}\cite{tang2022contrastive} proposed to improve point cloud segmentation performance by applying contrastive learning to point boundaries. Inspired by this approach, we propose a method of exploring CES in 3D semantic segmentation tasks by focusing on objects' boundaries: we first randomly sample anchors from the point cloud $\boldsymbol{\mathcal{P}}$, and then, within the neighbors of each anchor $\{\boldsymbol{X}_{a,i}\vert i=1,2,.., T\}$, choose embeddings with the same class label as positives $\{\boldsymbol{X}_{p,i}\vert i=1,2,.., T\}$, and those with different class labels as negatives $\{\boldsymbol{X}_{n,i}\vert i=1,2,..,T\}$. The triplets are constructed as $\{ \{\boldsymbol{X}_{a,i}, \boldsymbol{X}_{p,i}, \boldsymbol{X}_{n,i}\vert i=1,2,..,T\}\}$.

\subsubsection{CEM} \label{sec_cem} \cite{choy2019fully} presents embedding learning with triplet loss, and we demonstrate exploring CEM in 3D geometric feature learning tasks by following its sampling method. Specifically, given point clouds $\boldsymbol{\mathcal{P}}_i$ and $\boldsymbol{\mathcal{P}}_j$ and the relative transformation $\boldsymbol{\mathcal{T}}$, we first construct a list of all positive pairs from which we randomly draw $T$ pairs $\{(\boldsymbol{X}_{i_t,a}, \boldsymbol{X}_{j_t,a}), t=1,2,..,T\}$. Then we randomly choose the negative samples of each point as $\{(\boldsymbol{X}_{i_t,a}, \boldsymbol{X}_{i_t,n}), t=1,2,..,T\}$ and $\{(\boldsymbol{X}_{j_t,a}, \boldsymbol{X}_{j_t,n}), t=1,2,..,T\}$. Finally, The triplets are obtained as $\{(\boldsymbol{X}_{i_t,a}, \boldsymbol{X}_{j_t,a}, \boldsymbol{X}_{i_t,n}), t=1,2,..,T\}$ and $\{(\boldsymbol{X}_{j_t,a}, \boldsymbol{X}_{i_t,a}, \boldsymbol{X}_{j_t,n}), t=1,2,..,T\}$.


Sampling is a crucial component in making \sysname\ and \sysname+ possible for accurate uncertainty estimation in 3D dense prediction. Methods for uncertainty estimation in image recognition, such as \cite{warburg2021bayesian, zhang2021relative,chang2020data}, cannot be directly applied to 3D dense prediction as they are designed for global feature learning and not capable of handling large amounts of embeddings in 3D point clouds.
}
\section{Experimental Results}
\label{experiments}

\subsection{3D Geometric Feature Learning}


\textit{3D geometric feature learning} 
aims to train a deep neural network to map raw points in Euclidean space to a feature space, with the goal of having points with similar geometric characteristics be close to each other in the feature space. 
\revised{\cite{choy2019fully} studied different sampling strategies, including hardest-triplet sampling and random triplet sampling, and used triplet loss as the loss function. We build upon its model by exploring CEM with the sampling method described in Sec.\ref{sec_cem} and adapting their conventional triplet loss to our metric loss $L_M$.}


\begin{figure}[t]
    \centering
    \includegraphics[width=3in]{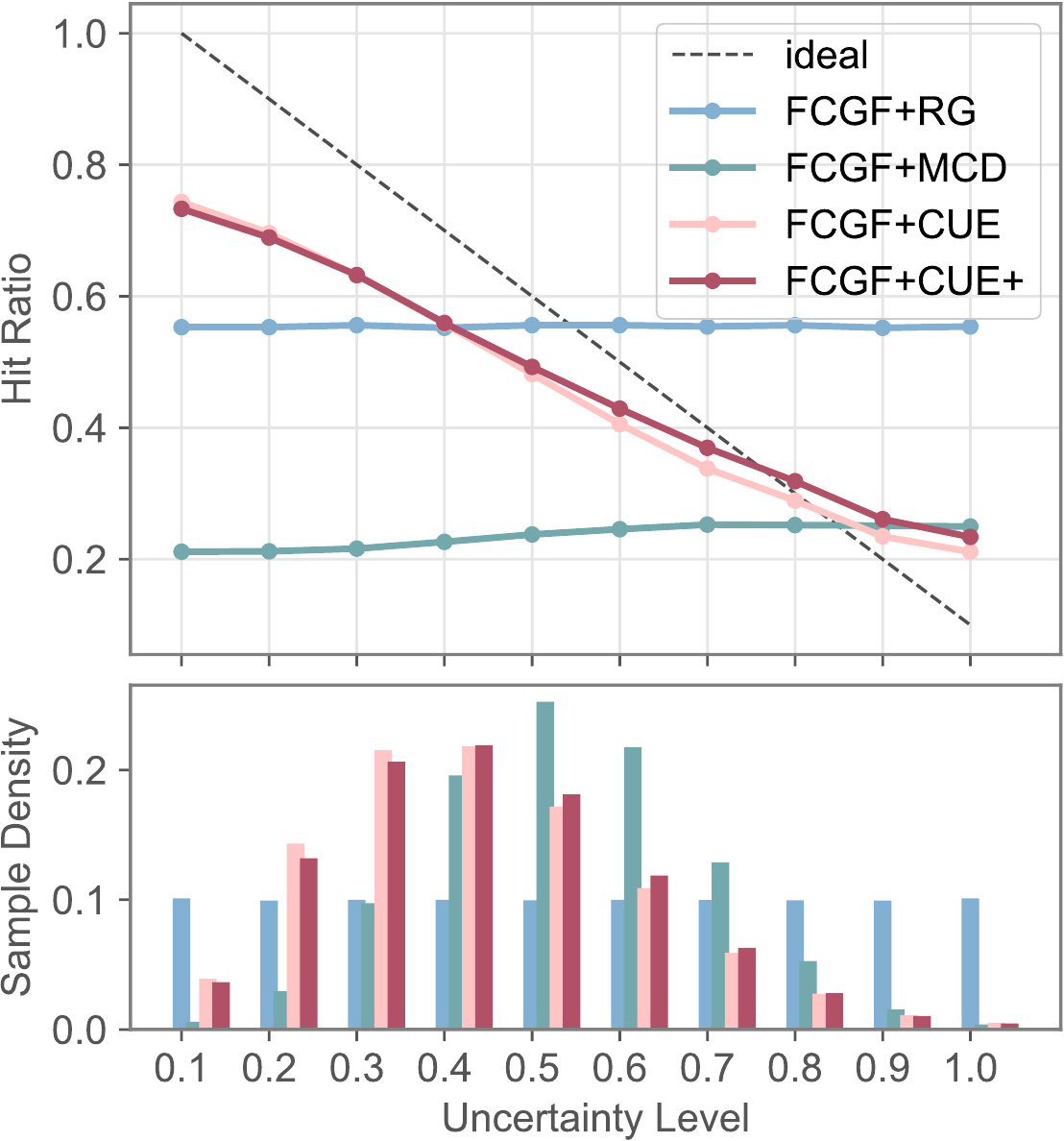} 
    \vspace{-0.3cm}
    \caption{Reliability diagram on the 3D Match Benchmark. \sysname\ and \sysname+ are closer to the ideally-calibrated line than others.} 
    \label{fig_ece_fcgf}
    \vspace{-0.2cm}
\end{figure}

\begin{table}
    \centering
    \caption{Predictive performance and uncertainty quality on the 3D Match Benchmark. 
    }    
    \label{table_fcgf}

    \begin{tabular}{l|c|c} 
    \hline
    Method                                 & FMR@0.05 ↑ & ECE ↓             \\ 
    \hline
    FPFH$^*$\cite{rusu2009fast}                                 & 36.4       & \textbackslash{}  \\
    PerfectMatch$^*$\cite{gojcic2019perfect}                          & 94.9       & \textbackslash{}  \\
    
    FCGF$^*$\cite{choy2019fully}                                  & 95.3       & \textbackslash{}  \\
    SpinNet\cite{ao2021spinnet}                                & 97.5       & \textbackslash{}  \\     
    FCGF\cite{choy2019fully}                               & 97.5       &     \textbackslash{}         \\    

    \hline
    FCGF+RG                               & 97.5       & 0.251             \\
    FCGF+MCD                             & 94.1       & 0.344             \\
    \rowcolor[rgb]{1,0.949,0.8} FCGF+\sysname  & 97.5       & 0.142             \\
    \rowcolor[rgb]{1,0.949,0.8} FCGF+\sysname+ & 97.6       & 0.135             \\
    \hline
    \end{tabular}
\begin{tablenotes}
    \footnotesize
    \item[1] $^*$ denotes predicting correspondences without a symmetric test \cite{horache20213d}.
\end{tablenotes}      
\vspace{-0.5cm}
\end{table}

\begin{figure}[htbp]
    \centering
    \includegraphics[width=3.3in]{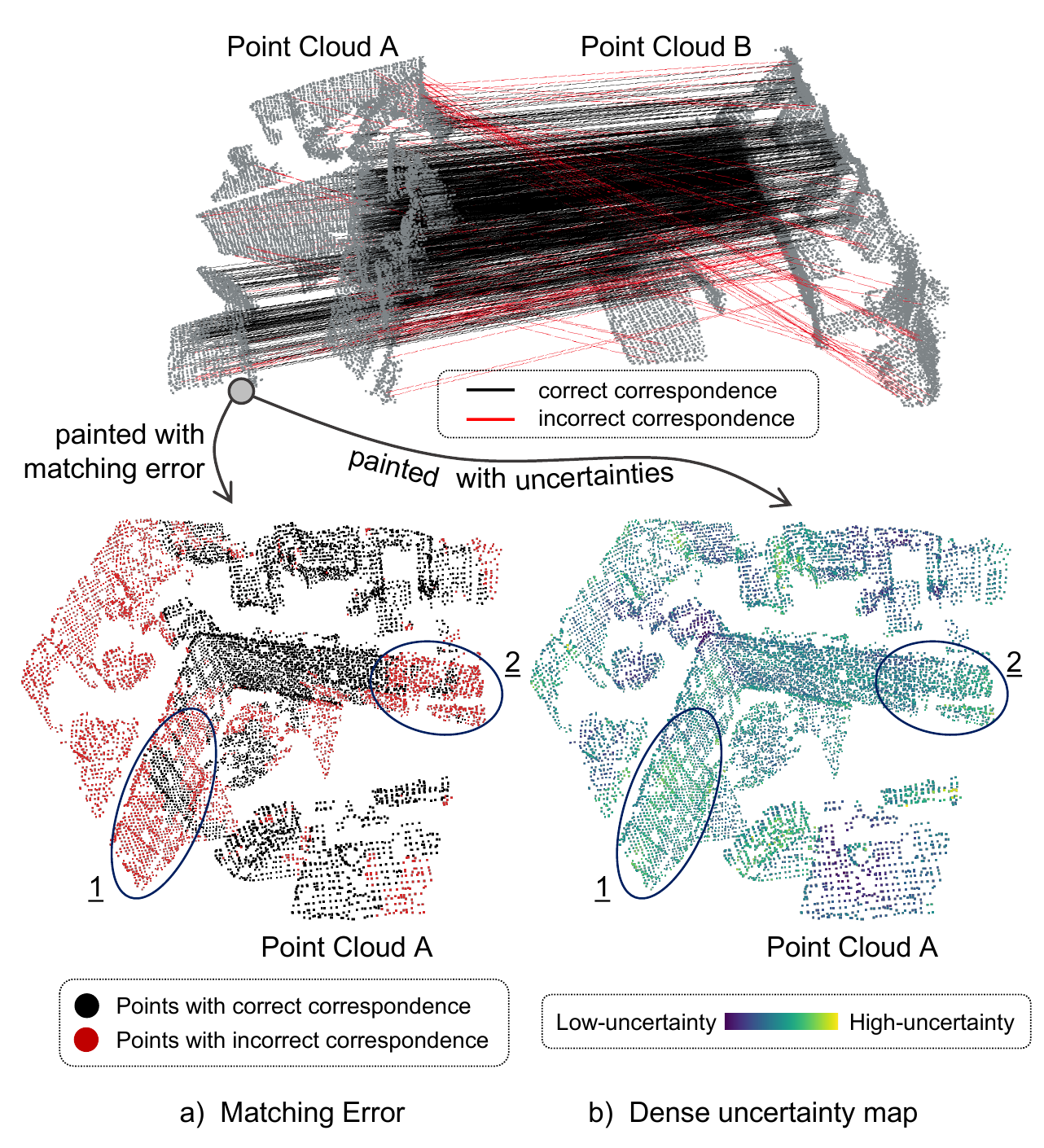} 
    \vspace{-0.3cm}
    \caption{Matching results and dense uncertainty map (estimated by \sysname+) of a point cloud from \rev{the} 3D Match Benchmark. Incorrect correspondences (area $1$ and $2$) tend to have high uncertainties.  } 
    \label{fig_qua_fcgf}
    \vspace{-0.6cm}
\end{figure}

\noindent \textbf{Datasets.} We use the 3D Match dataset \cite{zeng20173dmatch}, following the official training and evaluation splits.

\noindent \textbf{Model Architectures.} 
FCGF \cite{choy2019fully} is a 3D convolutional network that is the first to integrate metric learning in a fully-convolutional setting. We choose FCGF \cite{choy2019fully} as our backbone because it \rev{has} state-of-the-art predictive performance, fast training and inferencing. \rev{To add the ability to estimate the uncertainty of each point, the FCGF is integrated} with the proposed \sysname\ and \sysname+, as is shown in Fig. \ref{fig_network}.

\noindent \textbf{Training Details.}
We train FCGF following the original paper \cite{choy2019fully}, i.e., Hardest-contrastive loss, $100$ epochs with SGD optimizer and batch size $4$, learning rate starts from $0.1$ with exponential decay rate $0.99$, dada augmentation includes random scaling $\in [0.8, 1.2]$ and random rotation $\in [0 ^\circ , 360^\circ)$. 

\noindent \textbf{Competing Methods.}
\begin{itemize}
    \item Random Guess (RG): After training the FCGF, \revised{each point is assigned a random uncertainty value drawn from a uniform distribution}.
    \item MCD: We insert dropout layers with dropout rate $p=0.1$ after every convolutional layer. We take $N=40$ samples from the weights' posterior distribution at test time.
    \item \sysname: To \rev{maintain the} original predictive performance, we freeze the $\boldsymbol{\mu}$ branch and train $\boldsymbol{\Lambda}$ branches with $L_M$. 
    \item \sysname+: We freeze the $\boldsymbol{\mu}$ branch, and train $\boldsymbol{\Lambda}$ and $\boldsymbol{P}$ branches with $L_M$.
\end{itemize}
Note that MCD produces epistemic uncertainty, while our methods generate aleatoric uncertainty. 
\rev{MCD is included in this comparison for the sake of completeness.}

\noindent \textbf{Evaluation Metrics.}
To evaluate the predictive performance, we use Feature Matching Recall with $0.1m$ inlier distance threshold and $0.05$ inlier recall threshold (FMR@0.05) \cite{choy2019fully}. We adopt the widely used Expected Calibration Error (ECE) \cite{warburg2021bayesian} and the reliability diagram \cite{warburg2021bayesian} to evaluate uncertainty quality, where we calculate the Hit Ratio \cite{choy2019fully} of points in the same bin.

\noindent \textbf{Results.}
We evaluate the above methods on the 3D Match Benchmark \cite{zeng20173dmatch}. We establish correspondences by the nearest neighbor search in the embedding space, \rev{with} each correspondence \rev{having} an estimated uncertainty\footnote[1]{We follow the covariance formulation in \cite{warburg2021bayesian} and use the sum of two points' uncertainty as the correspondence's uncertainty.}. Table. \ref{table_fcgf} shows the predictive performance and uncertainty quality of different methods on the 3DMatch dataset. MCD shows degraded predictive performance due to the dropout layers negatively impacting the network's representation ability. \revised{Besides, the ECE of MCD is even worse than RG, meaning MCD fails to provide a sensible uncertainty.} Since the $\boldsymbol{\mu}$ branch is inherited from the backbone network, \sysname\ and \sysname+ do not sacrifice predictive accuracy. \revised{Compared with RG, \sysname\ reduces ECE by $43.4\%$. \sysname+ shows similar predictive performance as \sysname, but reduces the ECE of \sysname\ by $4.9\%$.}

\begin{figure}[htbp]
    \centering
    \includegraphics[width=3.3in]{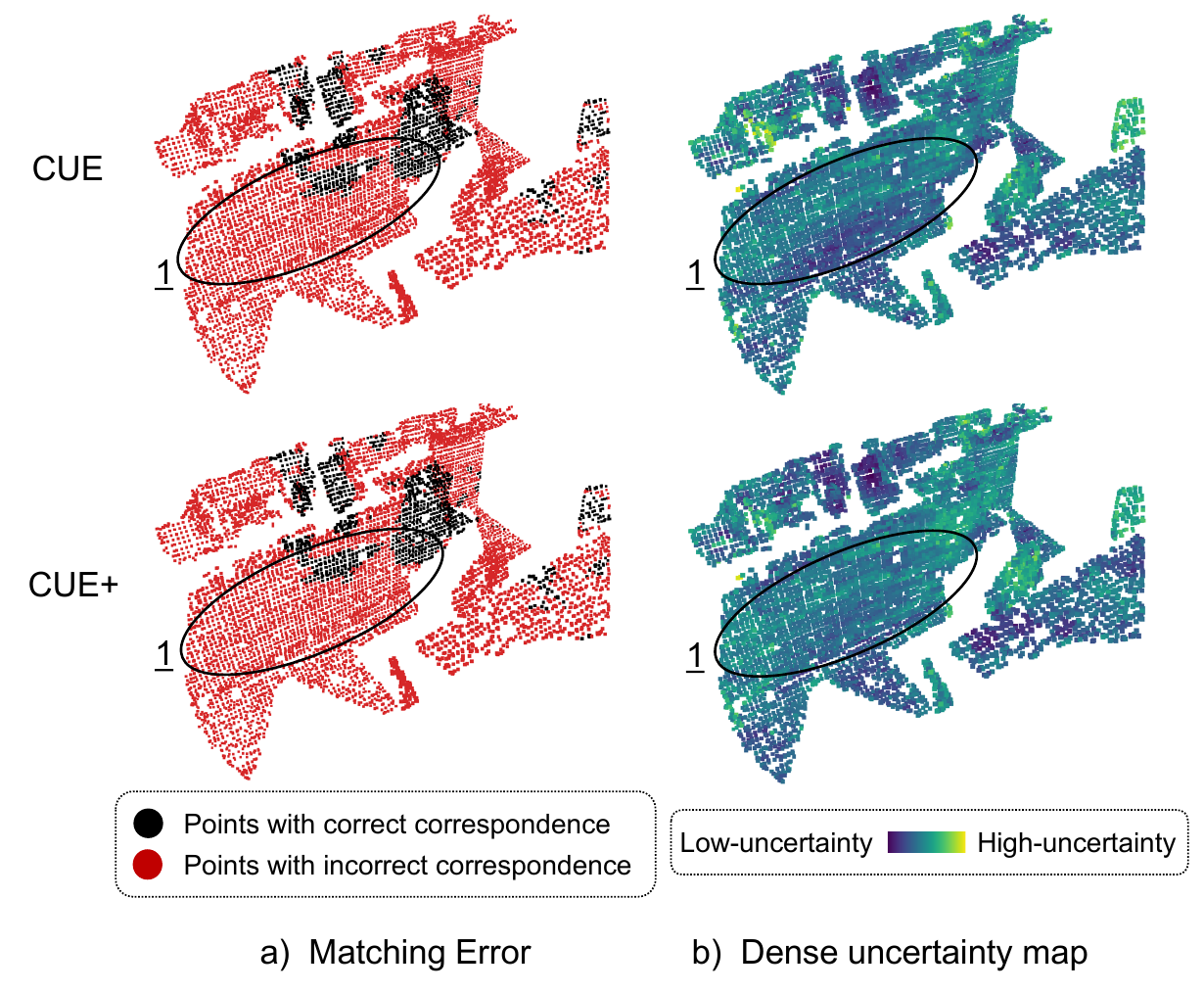} 
    \vspace{-0.3cm}
    \caption{\revised{Matching results and dense uncertainty map of a point cloud from the 3D Match Benchmark. The first row and second row denote results generated by \sysname\ and \sysname+, respectively. \sysname\ is overconfident about the incorrect correspondences (area $1$), while \sysname+ is properly confident.}} 
    \label{fig_qua_cuecuep}
    \vspace{-0cm}
\end{figure}

Fig. \ref{fig_qua_fcgf} shows the matching results and dense uncertainty map estimated by \sysname+ of a point cloud. We can observe that incorrect correspondences (area $1$ and $2$) tend to have high uncertainties. 
Fig. \ref{fig_ece_fcgf} presents the reliability diagram on the 3D Match Benchmark. The ideal line indicates that points with higher uncertainty levels should have lower hit ratios. \revised{It can be seen that the cuves of RG and MCD are  nearly flat, indicating them are not able to effectively associate uncertainty levels with hit ratios. 
The \sysname\ and \sysname+ are shown to have much closer lines to the ideal in the reliability diagram. In low-uncertainty regions (Uncertainty Level $\leq$ 0.4), the performance of \sysname\ and \sysname+ are similar, but  in high-uncertainty regions (0.5 $\leq$Uncertainty Level $\leq$ 0.7), \sysname\ is less effective than \sysname+ due to  overconfidence. The trend is shown in Fig. \ref{fig_qua_cuecuep}, where \sysname\ is overconfident about the incorrect correspondences ($1$ areas), while \sysname+ is properly confident.} 
\revised{
This demonstrates that the estimated uncertainty is highly practical, as it can be utilized as an effective tool for filtering incorrect correspondences when performing point cloud registration.
}

\revised{In summary, both the proposed \sysname\ and \sysname+  can estimated uncertainty for 3D geometric feature learning without compromising predictive performance, and \sysname+ shows better calibrated uncertainty than \sysname.}

\vspace{-0.2cm}
\subsection{3D Semantic Segmentation}

\begin{figure}[t]
    \centering
    \includegraphics[width=3in]{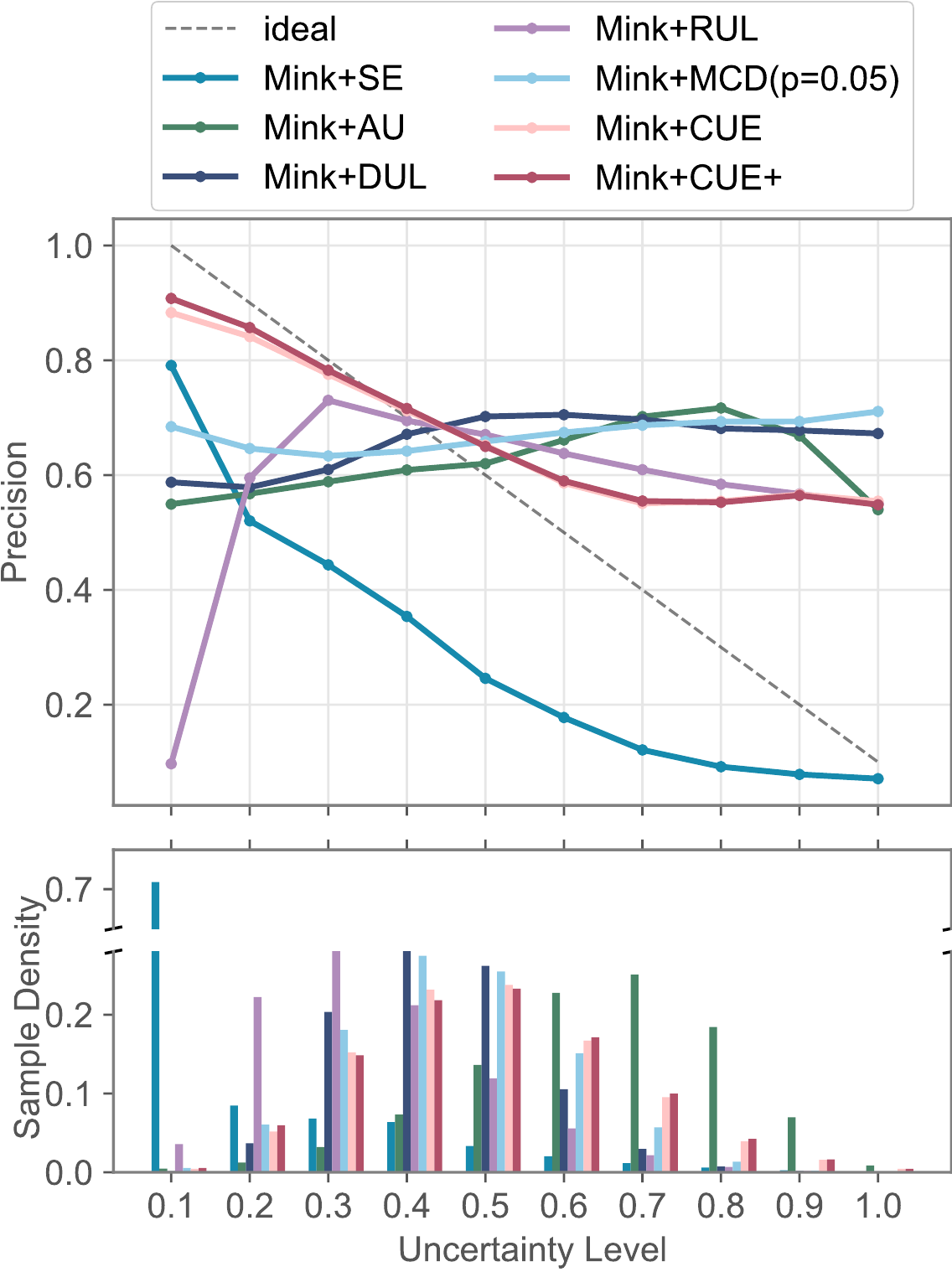} 
    \vspace{-0.3cm}
    \caption{Reliability diagram on the ScanNet validation split. \sysname\ and \sysname+ are closer to the ideally-calibrated line than other methods.} 
    \label{fig_ece_scannet}
    \vspace{-0.2cm}
\end{figure}

\textit{3D semantic segmentation} aims to learn a classification network that accurately predicts the class  of each point in a point cloud. We explore CES with sampling methods described in \ref{sec_ces}. 

\noindent \textbf{Datasets.} 
Following \cite{park2022fast}, we use the ScanNet dataset \cite{dai2017scannet} and evaluate models on the ScanNet validation split.

\noindent \textbf{Model Architectures.} 
We choose MinkowskiNet42 (Mink) \cite{choy20194d, park2022fast} as our 3D semantic segmentation backbone since it has high accuracy and low inference latency. The semantic segmentation network is the same as that in Fig. \ref{fig_network}, except that we add a convolution layer as the segmentation classifier before the L2-Normalization layer of the $\boldsymbol{\mu}$ branch. 

\begin{table}
    \centering
    \caption{Predictive performance and uncertainty quality on the ScanNet validation split. 
    \label{table_scannet}
    }        
    \begin{tabular}{l|l|c|c} 
        \hline
   & Method                             & mIOU ↑                              & ECE ↓                                \\ 
        \hline
        \multirow{5}{*}{\begin{tabular}[c]{@{}c@{}}\revised{Without} \\\revised{uncertainty} \\\revised{estimation} \end{tabular}} & PointNet\cite{qi2017pointnet}                           & 0.535                               & \textbackslash{}                     \\
                                                                                                  & PointConv\cite{wu2019pointconv}                          & 0.610                               & \textbackslash{}                     \\
                                                                                                  & KPConv
          deform \cite{thomas2019kpconv}                   & 0.692                               & \textbackslash{}                     \\
                                                                                                  & SparseConvNet\cite{graham20183d}                      & 0.693                               & \textbackslash{}                     \\
                                                                                                  & Mink\cite{choy20194d}                           & 0.715                               &   \textbackslash{}                              \\                                                        
        \hline
        \multirow{6}{*}{\begin{tabular}[c]{@{}c@{}}\revised{With}  \\\revised{uncertainty} \\\revised{estimation}\end{tabular}}   
        & Mink+SE\cite{jungo2019assessing}                           & 0.715                               & 0.251                                \\ 
        & Mink+AU\cite{kendall2017uncertainties}                          & 0.717                               & 0.254                                
        \\
        & \revised{Mink+DUL}\cite{chang2020data}                          & \revised{0.719}                               & \revised{0.173}                                \\      
        & \revised{Mink+RUL}\cite{zhang2021relative}                          & \revised{0.712}                               & \revised{0.187}                                \\            
                                                                                                  & Mink+MCD(p=0.20)                        & 0.658                               & 0.176                                \\
                                                                                                  & Mink+MCD(p=0.05)                        & 0.663                               & 0.170                                \\                                                                                                  & {\cellcolor[rgb]{1,0.949,0.8}}Mink+\sysname\  & {\cellcolor[rgb]{1,0.949,0.8}}0.721 & {\cellcolor[rgb]{1,0.949,0.8}}0.142  \\
                                                                                                  & {\cellcolor[rgb]{1,0.949,0.8}}Mink+\sysname+ & {\cellcolor[rgb]{1,0.949,0.8}}0.727 & {\cellcolor[rgb]{1,0.949,0.8}}0.141  \\
        \hline
        \end{tabular}
        \vspace{-0.6cm}
    \end{table}

\begin{figure}[htbp]
    \centering
    \includegraphics[width=3.4in]{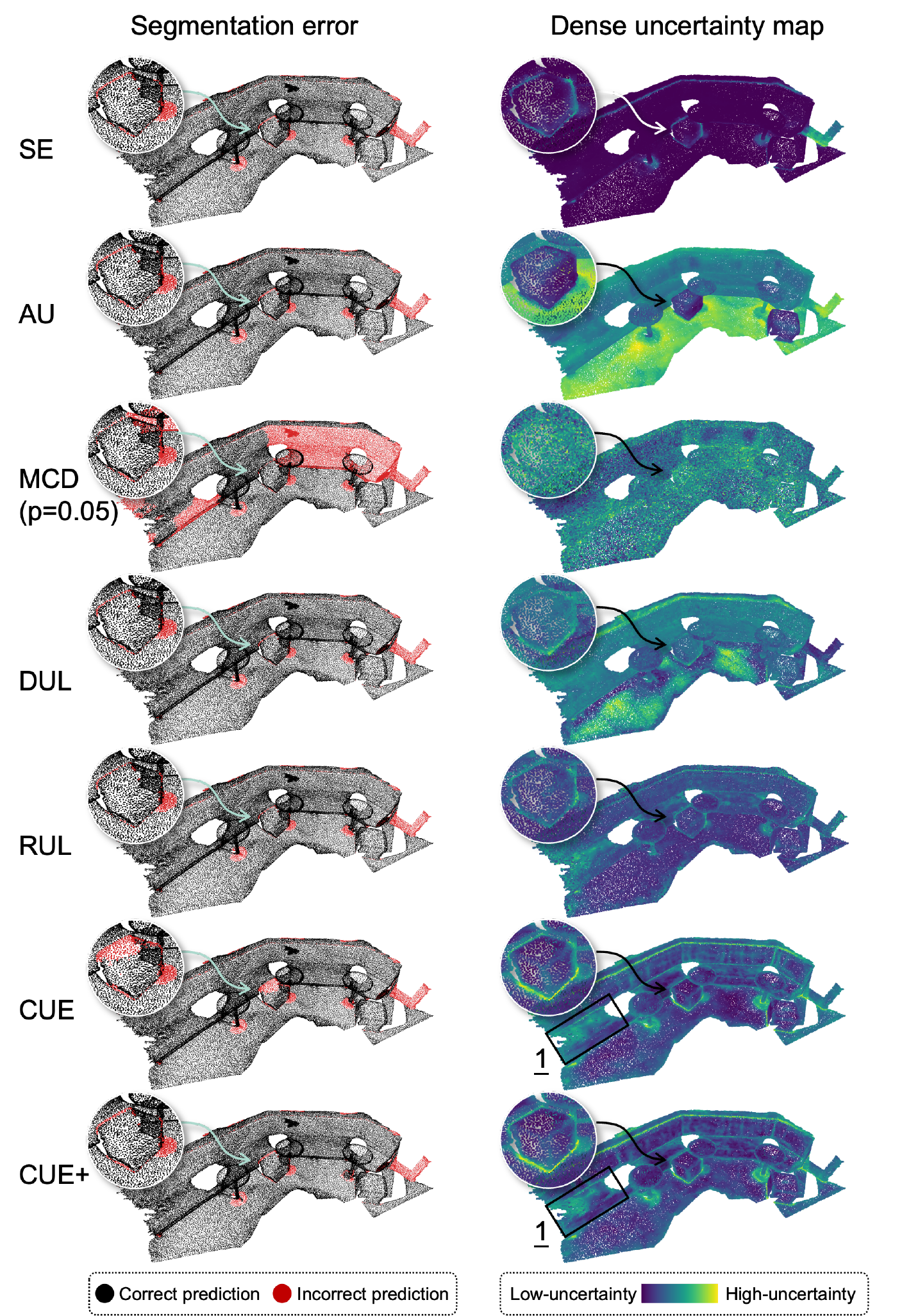} 
    \vspace{-0.2cm}
    \caption{Segmentation errors (left column) and dense uncertainty maps (right column) on a scene from the ScanNet validation split. \sysname\ and \sysname+ produce better-calibrated dense uncertainty maps than others. For correct predictions (rectangular area $1$),  \sysname\ is under-confident while \sysname+ is more confident than \sysname.} 
    \label{fig_qua_compare}
    \vspace{-0.4cm}
\end{figure}

\noindent \textbf{Training Details.}
We train the model for $10^5$ steps with an SGD optimizer, learning rate starting from $0.1$ with a cosine annealing schedule and a linear warmup. We use a batch size of $8$. More training details can be found in reference \cite{park2022fast}. 

\noindent \textbf{Competing Methods.}
We \rev{evaluate the performance of \sysname\ and \sysname+ against several well-known uncertainty estimation methods for image recognition or segmentation}:
\begin{itemize}
    \item Softmax Entropy \cite{jungo2019assessing} (SE): The uncertainty is calculated based on the entropy of softmax output as
    $ H = -\sum_c^Cp_c\log(p_c)/log(C) \in [0,1]$,
    where $C$ is the number of classes, $p_c$ is a probability by the softmax layer. 
    \item Aleatoric Uncertainty \cite{kendall2017uncertainties, jungo2019assessing} (AU): The Logit are modeled as a Gaussian distribution, with \rev{the} mean and \rev{the} variance predicted by two \rev{seperate} heads of the network. We use MC sampling (with $n=10$) to \rev{generate} samples from the logits distribution and optimize the network with \rev{the} cross entropy loss.
    \revised{\item Data Uncertainty Learning (DUL) \cite{chang2020data}: DUL is originally designed for face image recognition. We adapt it to our 3D semantic segmentation task by incorporating its {distributional representation} \cite{chang2020data} and replacing $L_M$ with its $\mathcal{L}_{cls}$.}
    \vspace{-0.4cm}
    \revised{\item Relative Uncertainty Learning (RUL) \cite{zhang2021relative}: RUL was proposed for facial expression recognition. We include its \textit{feature mixture} of \cite{zhang2021relative} and replace $L_M$ with its ${L}_{total}$.}
    \item MCD \cite{jungo2019assessing}: MCD estimates the epistemic uncertainty as enabling dropout at the test time approximates a random sampling of the model's weights. Test time inference is obtained by
    $ p_c = \frac{1}{N}\sum_n^Np_{n,c} $, where $p_c$ denotes the output of the Softmax layer. We set the number of MC samples $N=40$ as suggested by \cite{kendall2016modelling}. We evaluate MCD with two dropout probability settings: $p=0.2$ and $p=0.05$.  Since AU and MCD generates high-dimensional variance vectors, the variance vectors are transformed into uncertainty levels by $y(1-0.5q)+(1-y)(0.5q)$, where $q\in[0,1]$ is the normalized variance\cite{jungo2019assessing}.
    
    \item \sysname\ / \sysname+: We train the \sysname\ / \sysname+ network from scratch with a weighted sum of the cross entropy loss and the metric loss $L = L_{CE} + \lambda L_M$, 
    where we set $\lambda=1$ for all experiments.
    

\end{itemize}

\noindent \textbf{Evaluation Metrics.}
Mean Intersection over Union (mIoU) is a commonly used metric to evaluate the performance of image segmentation models. It is calculated as the ratio of the intersection of the ground-truth labels and predicted labels to their union. A higher mIoU indicates better performance. Additionally, the reliability diagram \cite{warburg2021bayesian} and ECE \cite{warburg2021bayesian} are \rev{used} to evaluate uncertainty quality, where we calculate the precision of \rev{the} points in each bin.

\noindent \textbf{Results.}
\revised{
Table. \ref{table_scannet} presents the predictive performance and uncertainty quality on the ScanNet validation split. As is shown in the `Without uncertainty estimation' section, Mink has the highest mIOU, indicating that it is the state-of-the-art 3D segmentation model. In the `With uncertainty estimation' section,  we observe that SE, AU, DUL, RUL and \sysname\ provide comparable predictive performance to Mink, while \sysname+ promotes Mink with the most significant boost of \revised{$0.012$} in mIoU. However, MCD shows degraded performance, which is attributed to the fact that dropout layers decrease the model's representative power. Regarding uncertainty quality, SE and AU show the highest ECE, which are $0.251$ and $0.254$, respectively. DUL, RUL and MCD indicate similar ECE results, which are $0.173$, $0.187$, and $0.176$(p=0.20)/$0.170$(p=0.05). In comparison, \sysname\ and \sysname+ provide significantly improved uncertainty with the ECE $0.142$ and $0.141$. This means \sysname+ reduces the ECE of the best existing method, MCD(p=0.05), by $16.5\%$. The results indicate that existing uncertainty estimation methods designed for image recognition (SE, AU, DUL and RUL) cannot produce satisfactory results on 3D dense prediction tasks as they fail to capture cross-point relations, particularly in a batch with massive points. Even though MCD shows relatively better uncertainty quality, this is achieved at the cost of predictive performance and processing time.
}

Fig. \ref{fig_ece_scannet} shows the reliability diagram on the ScanNet validation split. \revised{It can be seen that only SE, \sysname\ and \sysname+ show descending trends as the ideal line, while the other methods present opposite or inconsistent trends. However, SE deviates significantly from the ideal line, while \sysname\ is close to the ideal line and \sysname+ improves \sysname\ in the low-uncertainty region.} Fig. \ref{fig_qua_scannet} presents the qualitative results of \sysname+, where we can observe a significant correlation between segmentation prediction error and estimated uncertainty, i.e., Incorrect predictions tend to have high uncertainties. 

Fig. \ref{fig_qua_compare} presents segmentation errors and dense uncertainty maps by different methods on the ScanNet validation split. For incorrect predictions (black points in the magnified area), we can observe that SE fails to detect them and shows high confidence, while \sysname\ and \sysname+ are uncertain about those incorrect predictions. \revised{AU and DUL are under-confident in most areas, while RUL is over-confident in many points with incorrect predictions, e.g., the corners of the magnified areas and the table legs. And MCD (p=0.05) cannot produce sensible results.}
For correct predictions (Rectangular area $1$),
\sysname\ is under-confident while \sysname+ is more confident than \sysname.

\begin{table}
    \centering
    \caption{Computational Complexity}
    \label{table_computation}
    \begin{tabular}{l|c|c} 
    \hline
    Method                                & Num of Params ↓ & \begin{tabular}[c]{@{}c@{}}Processing time/\\Point cloud ↓\end{tabular}  \\ 
    \hline
    Mink\cite{choy20194d}                                  & 36.88M           & 0.163s                                                                   \\
    Mink+SE\cite{jungo2019assessing}                               & 36.88M          & 0.165s                                                                   \\
    Mink+AU\cite{kendall2017uncertainties}                               & 36.90M          & 0.172s                                                                   \\
    Mink+DUL\cite{chang2020data}                              & 36.90M          & 0.172s                                                                   \\
    Mink+RUL\cite{zhang2021relative}                              & 36.90M          & 0.165s                                                                   \\
    Mink+MCD                              & 36.88M          & 6.507s                                                                   \\
    \rowcolor[rgb]{1,0.945,0.8} Mink+CUE  & 36.90M          & 0.173s                                                                   \\
    \rowcolor[rgb]{1,0.945,0.8} Mink+CUE+ & 36.91M          & 0.174s                                                                   \\
    \hline
    \end{tabular}
    \vspace{-0.6cm}
    \end{table}
\revised{Table. \ref{table_computation} presents the computation complexity of the proposed \sysname/\sysname+ and comparing methods. We use NVIDIA A100 when evaluating networks' processing time during the inference phase. It is shown that Mink has the least trainable network parameters and the fastest processing speed. Although MCD does not add additional parameters to Mink, multiple forward propagations significantly increase MCD's processing time. AU, DUL, \sysname\ have more parameters since they have more layers, which brings ~$6\%$ overhead to Mink. Because \sysname+ has an additional branch, it has slightly more parameters and processing time compared to \sysname. Overall, \sysname\ and \sysname+ bring little overhead to the original network.}

The above results indicate that \sysname\ and \sysname+ provide better-calibrated uncertainty than existing methods without compromising predictive performance, and \sysname+ outperforms \sysname\ in both predictive performance and uncertainty quality with marginally more overhead. \revised{This demonstrates that explicitly expressing cross-point embedding interactions contributes to uncertainty estimation in 3D dense prediction tasks, where a low-rank multivariate Gaussian model is more effective than a diagonal one.}


\section{Conclusion}
\label{conclusion}
Observing the fact that dense prediction networks are sequential compositions of embedding learning networks and task-specific regressors (or classifiers), we propose \sysname\ that estimates uncertainty by building a probabilistic embedding model and enforcing metric alignments with a diagonal multivariate Gaussian model. We further propose \sysname+ that enhances cross-point interactions with a low-rank multivariate Gaussian model, which explicitly expresses off-diagonal elements' dependencies while maintaining computational efficiency. Experimental results on the 3D Match Benchmark and the ScanNet dataset have shown that \sysname\ and \sysname+ are generic and \revised{efficient (bring negligible overhead)} tools for uncertainty estimation in 3D dense prediction. \revised{Despite the promising results, there is still room for improvement. \sysname\//\sysname+ cannot guarantee correctly estimated uncertainties for \textit{all points} (see Fig. \ref{fig_qua_scannet}). We suppose that more sophisticated sampling strategies utilizing CES and CEM, such as using hardest triplet sampling or sampling among all points rather than points at objects' boundaries, would bring a better uncertainty estimation. This is an area of future research for us.}

\bibliographystyle{IEEEtran}
\bibliography{IEEEabrv, my_library}

\end{document}